\documentclass[letterpaper, 10 pt, journal, twoside]{IEEEtran}

\IEEEoverridecommandlockouts                              

\usepackage{amsmath,amssymb,amsfonts,booktabs,multirow,bm,balance,bbm,siunitx,hyperref,graphicx,algorithm}
\usepackage[noend]{algpseudocode}
\usepackage[usenames,dvipsnames]{color}
\usepackage{footmisc}
\usepackage[caption=false, font=footnotesize]{subfig}
\usepackage[noadjust]{cite}

\newcommand{\bs}{\boldsymbol}

\newcommand{\acc}{\ddot{\bs{s}}}
\newcommand{\vel}{\dot{\bs{s}}}
\newcommand{\pos}{\bs s}

\newcommand{\finalTime}{T}

\newcommand{\planePos}{\bs p}
\newcommand{\dir}{\bs n}
\newcommand{\dist}{d}
\newcommand{\extSetZ}{\mathcal{T}_{\mathrm{z}}}

\newcommand{\focalLength}{f}
\newcommand{\vehSphere}{\mathcal{S}}
\newcommand{\camFrame}{C}
\newcommand{\camSphere}{\mathcal{S}_\camFrame}
\newcommand{\radius}{r}
\newcommand{\distDotZ}{\dot{d}_z}
\newcommand{\camFrameX}{\bs x_\camFrame}
\newcommand{\camFrameY}{\bs y_\camFrame}
\newcommand{\camFrameZ}{\bs z_\camFrame}
\newcommand{\camPos}{\bs s_c}
\newcommand{\camVel}{\dot{\bs{s}}_c}
\newcommand{\intSet}{\mathcal{T}_{\mathrm{cross}}}
\newcommand{\intersectionTime}{t^\downarrow}

\newcommand{\bestCamPos}{\bs s_c^*}
\newcommand{\monoSects}{\mathcal{M}}
\newcommand{\monoSection}{\bs{\bar{s}}_c}
\newcommand{\deepestPoint}{\bs{\bar{s}}}
\newcommand{\pyramid}{\mathcal{P}}
\newcommand{\pyramidSet}{\mathcal{G}}
\newcommand{\depthImage}{\mathcal{D}}

\newcommand{\pixel}{p}
\newcommand{\expDir}{\bs d}

\newcommand{\occSpace}{\mathcal{O}}
\newcommand{\freeSpace}{\mathcal{F}}
\newcommand{\unknownSpace}{\mathcal{U}}
\newcommand{\unknownDist}{l}
\newcommand{\pyramidExp}{\mathcal{P}_{exp}}

\newcommand{\numPixels}{n}

\author{Nathan Bucki$^{1}$, Junseok Lee$^{1}$, and Mark W. Mueller$^{1}$%
	\thanks{Manuscript received: February 24, 2020; Revised May 15, 2020; Accepted June 11, 2020.}
	\thanks{This paper was recommended for publication by Editor-in-Chief Allison Okamura and Editor Tamim Asfour upon evaluation of the Associate Editor and Reviewers' comments.
		This work was supported by the Berkeley Fellowship for Graduate Study, the Berkeley DeepDrive project `Autonomous Aerial Robots in Dense Urban Environments', and the China High-Speed Railway Technology Co., Ltd.} 
	\thanks{$^{1}$The authors are with the High Performance Robotics Lab, University of California, Berkeley, CA, USA.
		{\tt\footnotesize nathan\_bucki@berkeley.edu, junseok\_lee@berkeley.edu, mwm@berkeley.edu}}%
	\thanks{Digital Object Identifier (DOI): see top of this page.}
}

\title{
Rectangular Pyramid Partitioning using Integrated Depth Sensors (RAPPIDS): A Fast Planner for Multicopter Navigation
}

\begin{document}

\maketitle

\begin{abstract}
We present RAPPIDS: a novel collision checking and planning algorithm for multicopters that is capable of quickly finding local collision-free trajectories given a single depth image from an onboard camera.
The primary contribution of this work is a new pyramid-based spatial partitioning method that enables rapid collision detection between candidate trajectories and the environment.
By leveraging the efficiency of our collision checking method, we shown how a local planning algorithm can be run at high rates on computationally constrained hardware, evaluating thousands of candidate trajectories in milliseconds.
The performance of the algorithm is compared to existing collision checking methods in simulation, showing our method to be capable of evaluating orders of magnitude more trajectories per second.
Experimental results are presented showing a quadcopter quickly navigating a previously unseen cluttered environment by running the algorithm on an ODROID-XU4 at 30 Hz.
\end{abstract}
\begin{IEEEkeywords}
	Reactive and Sensor-Based Planning, Collision Avoidance, Aerial Systems: Perception and Autonomy, Motion and Path Planning
\end{IEEEkeywords}

\section{Introduction} \label{sec:intro}

\IEEEPARstart{T}{he} ability to perform high-speed flight in cluttered, unknown environments can enable a number of useful tasks, such as the navigation of a vehicle through previously unseen areas and rapid mapping of new environments.
Many existing planning algorithms for navigation in unknown environments have been developed for multicopters, and can generally be classified as map-based algorithms, memoryless algorithms, or a mixture of the two.

Map-based algorithms first fuse sensor data into a map of the surrounding environment, and then perform path planning and collision checking using the stored map.
For example, \cite{oleynikova2016continuous} uses a local map to solve a nonconvex optimization problem that returns a smooth trajectory which remains far from obstacles.
Similarly, \cite{gao2017gradient, liu2016high, chen2016online, tordesillas2019fastrap} each find a series of convex regions in the free-space of a dynamically updated map, and then use optimization-based methods to find a series of trajectories through the convex regions.
Although these methods are generally able to avoid getting stuck in local minima (e.g. a dead end at the end of a hallway), they generally require long computation times to fuse recent sensor data into the global map.

Memoryless algorithms, however, only use the latest sensor measurements for planning.
For example, \cite{florence2016integrated} and \cite{lopez2017aggressive} both use depth images to perform local planning by organizing the most recently received depth data into a k-d tree, which enables the distance from a given point to the nearest obstacle to be quickly computed.
The k-d tree is then used to perform collision checking on a number of candidate trajectories, at which point the optimal collision-free candidate trajectory is chosen to track.
A different memoryless algorithm is presented in \cite{matthies2014stereo} which inflates obstacles in the depth image based on the size of the vehicle, allowing for trajectories to be evaluated directly in image space.
In \cite{zhang2019maximum}, a significant portion of computation is performed offline in order to speed up online collision checking.
The space around the vehicle is first split into a grid, a finite set of candidate trajectories are generated, and the grid cells with which each trajectory collides are then computed.
Thus, when flying the vehicle, if an obstacle is detected in a given grid cell, the corresponding trajectories can be quickly determined to be in collision.
However, such offline methods have the disadvantage of constraining the vehicle to a less expressive set of possible candidate trajectories, e.g. forcing the vehicle to only travel at a single speed.

Several algorithms also leverage previously gathered data while handling the latest sensor measurements separately, allowing for more collision-free trajectories to be found than when using memoryless methods while maintaining a fast planning rate.
For example, in \cite{barry2015pushbroom} a stereo camera pair is used onboard a fixed-wing UAV to detect obstacles at a specific distance in front of the vehicle, allowing for a local map of obstacles to be generated as the vehicle moves forward.
In \cite{florence2018nanomap} a number of recent depth images are used to find the minimum-uncertainty view of a queried point in space, essentially giving the vehicle a wider field of view for planning.
Additionally, in \cite{ryll2019efficient} the most recent depth image is both organized into a k-d tree and fused into a local map, allowing for rapid local planning in conjunction with slower global planning.

Although the previously discussed planning algorithms have been shown to perform well in complex environments, they typically require the use of an onboard computer with processing power roughly equivalent to a modern laptop.
This requirement can significantly increase the cost, weight, and power consumption of a vehicle compared to one with more limited computational resources.
We address this problem by introducing a novel spatial partitioning and collision checking method to find collision-free trajectories through the environment at low computational cost, enabling rapid path planning on vehicles with significantly lower computational resources than previously developed systems.

The proposed planning algorithm, classified as a memoryless algorithm, takes the latest vehicle state estimate and a single depth image from an onboard camera as input.
The depth image is used to generate a number of rectangular pyramids that approximate the free space in the environment.
As described in later sections, the use of pyramids in conjunction with a continuous-time representation of the vehicle trajectory allows for any given trajectory to be efficiently labeled as either remaining in the free space, i.e. inside the generated pyramids, or as being potentially in collision with the environment.
Thus, a large number of candidate trajectories can be quickly generated and checked for collisions, allowing for the lowest cost trajectory, as specified by a user provided cost function, to be chosen for tracking until the next depth image is captured.
Furthermore, by choosing a continuous-time representation of the candidate trajectories, each trajectory can be quickly checked for input feasibility using existing methods.

The use of pyramids to approximate the free space is advantageous because they can be created efficiently by exploiting the structure of a depth image, they can be generated on an as-needed basis (avoiding the up-front computation cost associated with other spatial partitioning data structures such as k-d trees), and because they inherently prevent occluded/unknown space from being marked as free space.
Additionally, because our method is a memoryless method rather than a map-based method, it is robust to common mapping errors resulting from poor state estimation (e.g. incorrect loop closures).

\section{System Model and Relevant Properties} \label{sec:background}

In this section we describe the form of the trajectories used for planning and several of their relevant properties.
These properties are later exploited to perform efficient collision checking between the trajectories and the environment.

We assume the vehicle is equipped with sensors capable of producing depth images that can be modeled using the standard pinhole camera model with focal length $\focalLength$.
Let the depth-camera-fixed frame $\camFrame$ be located at the focal point of the image with z-axis $\camFrameZ$ perpendicular to the image plane.
The point at position $(X,Y,Z)$ written in the depth-camera-fixed frame is then located $x = \focalLength\frac{X}{Z}$ pixels horizontally and $y = \focalLength\frac{Y}{Z}$ pixels vertically from the image center with depth value $Z$.

\subsection{Trajectory and Collision Model}
We follow \cite{mueller2015computationally} and \cite{bucki2019rapid} in modeling the desired position trajectory of the center of mass of the vehicle as the minimum jerk trajectory between two states, which corresponds to a fifth order polynomial in time.
Let $\pos(t)$, $\vel(t)$, and $\acc(t) \in \mathbb{R}^3$ denote the position, velocity, and acceleration of the center of mass of the vehicle relative to a fixed point in an inertial frame.
The desired position trajectory is then defined by the initial state of the vehicle, defined by $\pos(0)$, $\vel(0)$, and $\acc(0)$, the duration of the trajectory $\finalTime$, and the desired state of the vehicle at the end of the trajectory, defined by $\pos(\finalTime)$, $\vel(\finalTime)$, and $\acc(\finalTime)$.
A desired position trajectory of the vehicle can then be written as
\begin{equation}\label{eq:trajDef}
\pos(t) = \frac{\bs \alpha}{120}t^5 + \frac{\bs \beta}{24}t^4 + \frac{\bs \gamma}{6}t^3 + \frac{\acc(0)}{2}t^2 + \vel(0)t + \pos(0)
\end{equation}
where $\bs \alpha$, $\bs \beta$, and $\bs \gamma \in \mathbb{R}^3$ are constants that depend only on $\pos(\finalTime)$, $\vel(\finalTime)$, $\acc(\finalTime)$, and $\finalTime$.
Note that the thrust direction of the vehicle (and thus its roll and pitch) is defined by its acceleration $\acc(t)$.
We refer the reader to \cite{mueller2015computationally} for details regarding this relation, as well as methods for quickly checking whether constraints on the minimum and maximum thrust and magnitude of the angular velocity of the multicopter are satisfied throughout the duration of the trajectory.
We define $\pos(t)$ as a collision-free trajectory if a sphere $\vehSphere$ centered at $\pos(t)$ that contains the vehicle does not intersect with any obstacles for the duration of the trajectory.

We additionally define a similar trajectory $\camPos(t)$ with initial position $\camPos(0)$ coincident with the depth-camera-fixed frame $\camFrame$ such that
\begin{equation}\label{eq:camTrajDef}
\camPos(t) = \frac{\bs \alpha}{120}t^5 + \frac{\bs \beta}{24}t^4 + \frac{\bs \gamma}{6}t^3 + \frac{\acc(0)}{2}t^2 + \vel(0)t + \camPos(0)
\end{equation}

The trajectory $\camPos(t)$ is used for collision checking rather than directly using the trajectory of the center of mass $\pos(t)$ because $\camPos(t)$ originates at the focal point of the depth image, allowing for the use of the advantageous properties of $\camPos(t)$ described in the following subsections.
Let $\camSphere$ be a sphere centered at $\camPos(t)$ with radius $\radius$ that contains the sphere $\vehSphere$.
If the larger sphere $\camSphere$ does not intersect with any obstacles for the duration of the trajectory, we can then also verify that the sphere containing the vehicle $\vehSphere$ remains collision-free.
Thus, we can use $\camPos(t)$ and its advantageous properties to check if $\pos(t)$ is collision-free at the expense of a small amount of conservativeness related to the difference in size between the outer sphere $\camSphere$ and inner sphere $\vehSphere$.

\subsection{Trajectory sections with monotonically changing depth}\label{sec:monotonic}
We split a given trajectory, e.g. $\camPos(t)$, into different sections with monotonically increasing or decreasing distance along the z-axis $\camFrameZ$ of the depth-camera-fixed frame $C$ (i.e. into the depth image) as follows.
First, we compute the rate of change of $\camPos(t)$ along $\camFrameZ$ as $\distDotZ(t) = \camFrameZ \cdot \camVel(t)$.
Then, by solving $\distDotZ(t) = 0$ for $t \in [0,\finalTime]$, we can find points $\extSetZ$ at which $\camPos(t)$ might change direction along $\camFrameZ$, defined as
\begin{equation}\label{eq:extSetZ}
\extSetZ = \{t_i : t_i \in [0, \finalTime],\; \distDotZ(t_i) = 0\} \cup \{0, \finalTime\}
\end{equation}
Note $\extSetZ$ can be computed in closed-form because it only requires finding the roots of the fourth order polynomial $\distDotZ(t) = 0$.

Splitting the trajectory into these monotonic sections is advantageous for collision checking because we can compute the point of each monotonic section with the deepest depth from the camera by simply evaluating the endpoints of the section.
Thus, we can avoid performing collision checking with any obstacles at a deeper depth than the deepest point of the trajectory.

\subsection{Trajectory-Plane Intersections}\label{sec:intersections}
A similar method can be used to quickly determine if and/or when a given trajectory intersects with an arbitrary plane defined by a point $\planePos \in \mathbb{R}^3$ and unit normal $\dir \in \mathbb{R}^3$.
Let the distance of the trajectory from the plane be written as $\dist(t) = \dir \cdot (\camPos(t) - \planePos)$.
The set of times $\intSet$ at which $\camPos(t)$ intersects the given plane are then defined as
\begin{equation}
\intSet = \{t_i : t_i \in [0, \finalTime],\; \dist(t_i) = 0\}
\end{equation}
requiring the solution of the equation $\dist(t) = 0$.
Unfortunately, $\dist(t)$ is in general a fifth order polynomial, meaning that its roots cannot be computed in closed-form and require more computationally expensive methods to find.

To this end, we extend \cite{mueller2015computationally} and \cite{bucki2019rapid} in presenting the conditions under which finding $\intSet$ can be reduced to finding the roots of a fourth order polynomial.
Specifically, if a single crossing time of $\dist(t)$ is known a priori, $\dist(t) = 0$ can be solved by factoring out the known root and solving the remaining fourth order polynomial.
This property is satisfied, for example, by any plane with $\planePos := \camPos(0)$ (i.e. a plane that intersects the initial position of the trajectory), resulting in the following equation for $\dist(t)$:
\begin{equation}\label{eq:distFactored}
\dist(t) = \dir \cdot \left(\frac{\bs \alpha}{120}t^4 + \frac{\bs \beta}{24}t^3 + \frac{\bs \gamma}{6}t^2 + \frac{\acc(0)}{2}t + \vel(0)\right)t
\end{equation}

Thus, the remaining four unknown roots of \eqref{eq:distFactored} can be computed using the closed-form equation for the roots of a fourth order polynomial, allowing for $\intSet$ to be computed extremely quickly.
As described in the following section, we exploit this property in order to quickly determine when a given trajectory leaves a given pyramid.

\section{Algorithm Description} \label{sec:algDescription}

In this section we describe the our novel pyramid-based spatial partitioning method, its use in performing efficient collision checking, and the algorithm used to search for the best collision-free trajectory.\footnote{An implementation of the algorithm can be found at \url{https://github.com/nlbucki/RAPPIDS}}

\subsection{Pyramid generation}
\begin{figure}
	\includegraphics[width=\columnwidth]{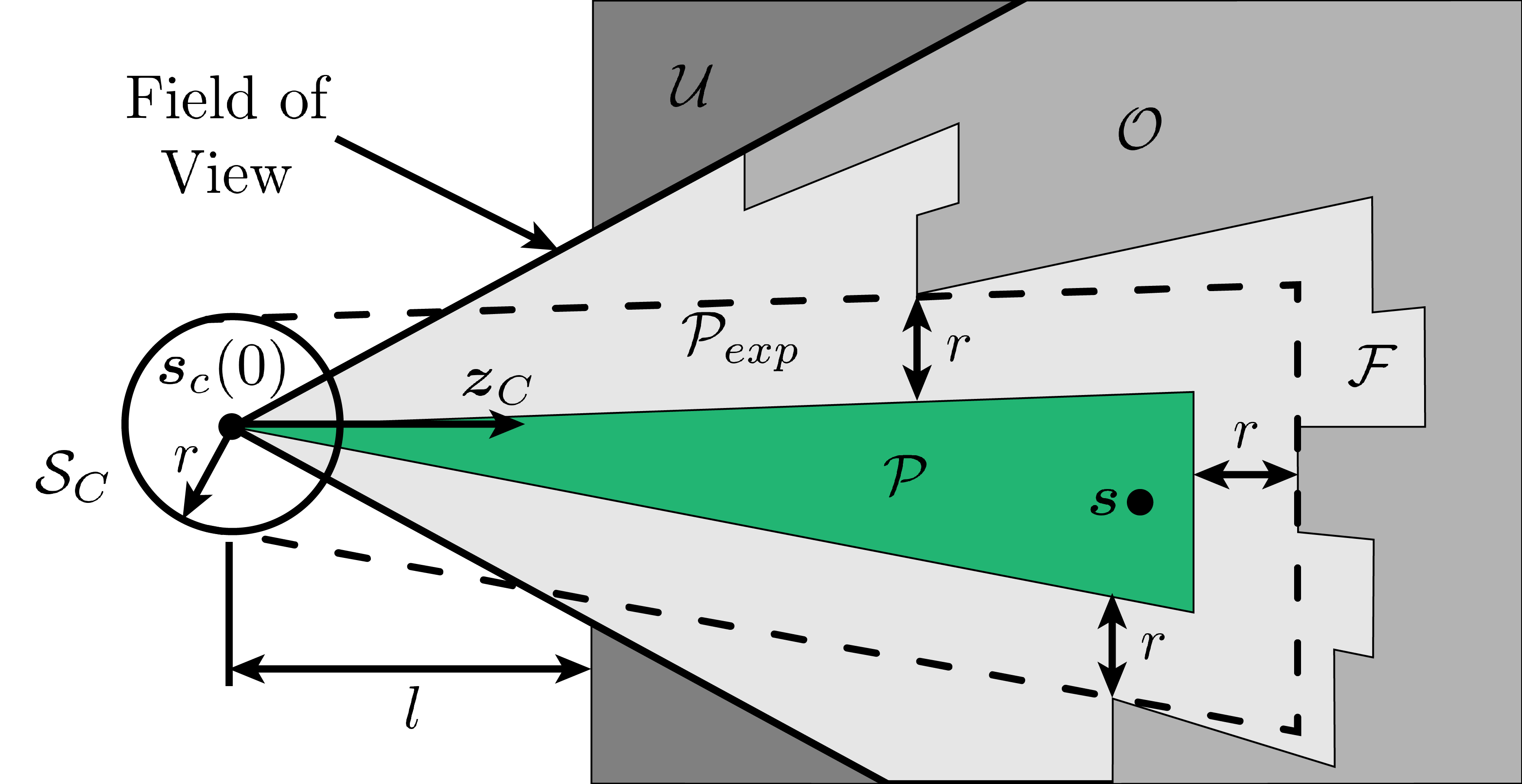}
	\centering
	\caption{2D side view illustrating the generation of a single pyramid $\pyramid$, shown in green, from a single depth image and given point $\bs s$. The depth values of each pixel are used to define the division between free space $\freeSpace$ and occupied space $\occSpace$. Because the depth camera has a limited field of view, we additionally consider any space outside the field of view farther than distance $\unknownDist$ from the camera to be unknown space $\unknownSpace$, which is treated the same as occupied space. The expanded pyramid $\pyramidExp$ (dash-dotted line) is first generated such that it does not contain any portion of $\occSpace$ or $\unknownSpace$, and then used to define pyramid $\pyramid$ such that it is distance $\radius$ from any obstacles.}
	\label{fig:pyramid}
\end{figure}

For each depth image generated by the vehicle's depth sensor, we partition the free space of the environment using a number of rectangular pyramids, where the apex of each pyramid is located at the origin of the depth camera-fixed frame $\camFrame$ (i.e. at $\camPos(0)$), and the base of each pyramid is located at some depth $Z$ and is perpendicular to the z-axis of the depth camera-fixed frame $\camFrameZ$ as shown in Figure \ref{fig:pyramid}.

The depth value stored in each pixel of the image is used to define the separation of free space $\freeSpace$ and occupied space $\occSpace$.
We additionally treat the space $\unknownSpace$ located outside the field of view of the camera at depth $\unknownDist$ from the camera focal point as occupied space.
Pyramid $\pyramid$ is defined such that while trajectory $\camPos(t)$ remains inside $\pyramid$, the sphere containing the vehicle $\camSphere$ will not intersect with any occupied space, meaning that the segment of $\camPos(t)$ inside $\pyramid$ can be considered collision-free.

Note that if $\camPos(t)$ remains inside the pyramid, we can not only guarantee that the vehicle will not collide with any obstacles detected by the depth camera, but that the vehicle will not collide with any occluded obstacles either.
This differs from other methods that treat each pixel in the depth image as an individual obstacle to be avoided, which can result in the generation of over-optimistic trajectories that may collide with unseen obstacles.
Furthermore, our method straightforwardly incorporates field of view constraints, allowing for the avoidance of nearly all unseen obstacles in addition to those detected by the depth camera.

The function $\textproc{InflatePyramid}$ is used to generate a pyramid $\pyramid$ by taking an initial point $\bs s$ as input and returning either a pyramid containing $\bs s$ or a failure indicator.
In this work we choose $\bs s$ to be the endpoint of a given trajectory that we wish to check for collisions, and only generate a new pyramid if $\bs s$ is not already contained in an existing pyramid (details are provided in the following subsection).
We start by projecting $\bs s$ into the depth image and finding the nearest pixel $\pixel$.
Then, pixels of the image are read in a spiral about pixel $\pixel$ in order to compute the largest possible expanded rectangle $\pyramidExp$ that does not contain any occupied space.
Finally, pyramid $\pyramid$ is computed by shrinking the expanded pyramid $\pyramidExp$ such that each face of $\pyramid$ is not within vehicle radius $\radius$ of occupied space.
Further details regarding our implementation of $\textproc{InflatePyramid}$ can be found online.\footnotemark[\value{footnote}]

This method additionally allows for pyramid generation failure indicators to be returned extremely quickly.
For example, consider the case where the initial point $\bs s$ exists inside occupied space $\occSpace$.
Then, only the depth value of the nearest pixel $\pixel$ must be read before finding that no pyramid containing $\bs s$ can be generated, requiring only a single pixel of the depth image to be processed.
This property greatly reduces the number of operations required to determine when a given point is in collision with the environment.

\subsection{Collision checking using pyramids}
\begin{figure}
	\includegraphics[width=\columnwidth]{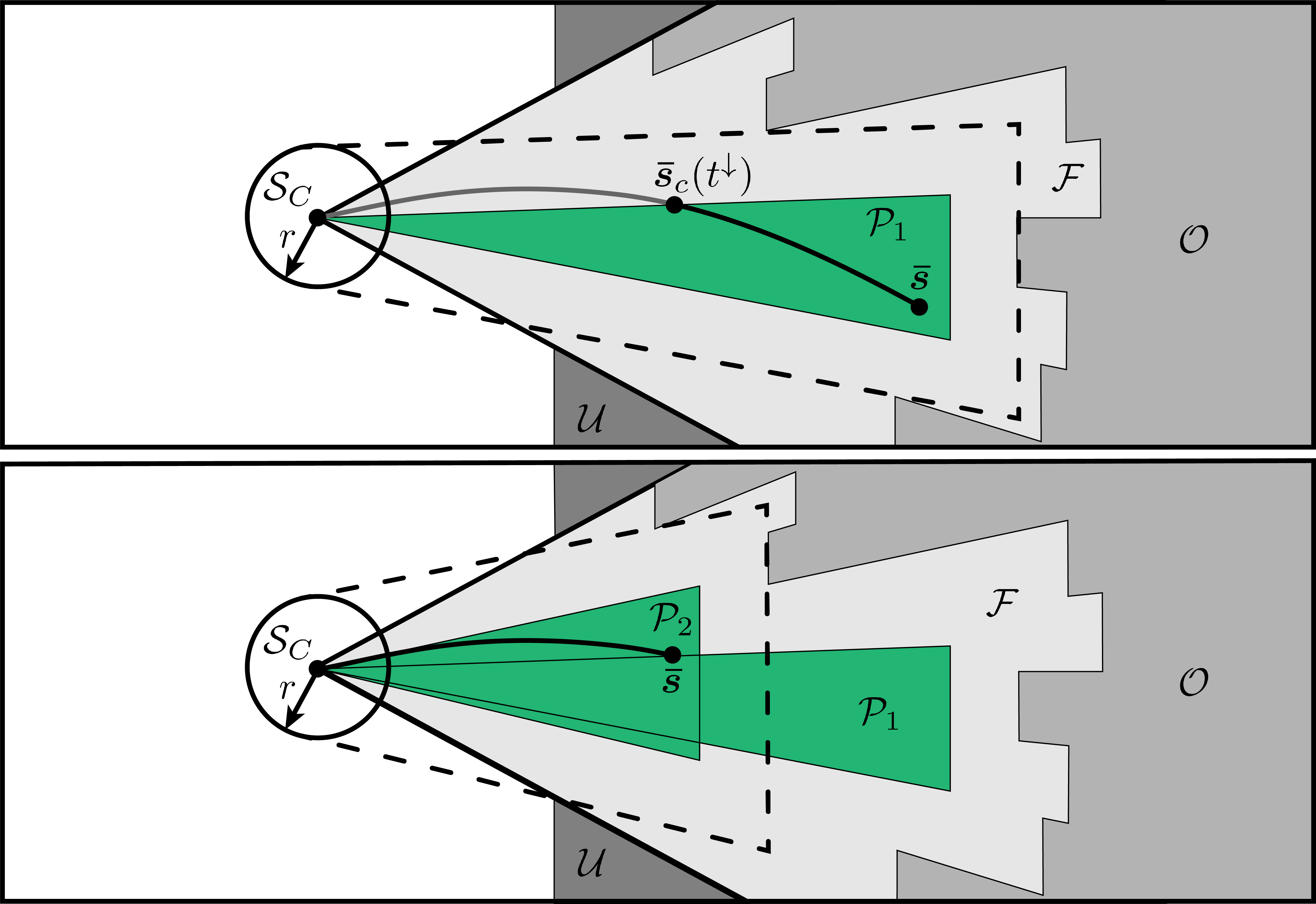}
	\centering
	\caption{2D example of the collision checking method described by Algorithm \ref{algo:collisionCheck} as used to check a given trajectory for collisions. The trajectory is first split into sections with monotonically changing depth, which are stored in list $\monoSects$. Top: A single trajectory section $\monoSection(t)$ is chosen from list $\monoSects$. The deepest point of the trajectory section $\deepestPoint$ is computed and pyramid $\pyramid_1$ containing $\deepestPoint$ is generated. The trajectory  $\monoSection(t)$ is then subdivided into a section that remains inside the pyramid (black) and a section that leaves the pyramid (gray). Bottom: The trajectory section that leaves $\pyramid_1$ is checked for collisions in the same manner. Pyramid $\pyramid_2$ is generated using the deepest point of the trajectory section, and then used to verify that the trajectory section does not collide with the environment.}
	\label{fig:collisionCheck}
\end{figure}

Algorithm \ref{algo:collisionCheck} describes how the set of all previously generated pyramids $\pyramidSet$ is used to determine whether a given trajectory $\camPos(t)$ will collide with the environment.
A trajectory is considered collision-free if it remains inside the space covered by $\pyramidSet$ for the full duration of the trajectory.
An example of Algorithm \ref{algo:collisionCheck} is shown in Figure \ref{fig:collisionCheck}.

\begin{algorithm}
	\caption{Single Trajectory Collision Checking}
	\label{algo:collisionCheck}
	\begin{algorithmic}[1]
		\State \textbf{input:} Trajectory $\camPos(t)$ to be checked for collisions, set of all previously found pyramids $\pyramidSet$, depth image $\depthImage$
		\State \textbf{output:} Boolean indicating if trajectory is collision-free, updated set of pyramids $\pyramidSet$
		
		\Function{IsCollisionFree}{$\camPos(t)$, $\pyramidSet$, $\depthImage$}
		\State $\monoSects \gets \textproc{GetMonotonicSections}(\camPos(t))$
		\While{$\monoSects$ is not empty}
		\State $\monoSection(t) \gets \textproc{pop}(\monoSects)$
		\State $\deepestPoint \gets \textproc{DeepestPoint}(\monoSection(t))$
		\State $\pyramid \gets \textproc{FindContainingPyramid}(\pyramidSet, \deepestPoint)$
		\If{$\pyramid$ is null}
		\State $\pyramid \gets \textproc{InflatePyramid}(\deepestPoint, \depthImage)$
		\If{$\pyramid$ is null}
		\State \textbf{return} false
		\EndIf
		\State $\textproc{push}(\pyramid) \rightarrow \pyramidSet$
		\EndIf
		\State $\intersectionTime \gets \textproc{FindDeepestCollisionTime}(\pyramid, \monoSection(t))$
		\If{$\intersectionTime$ is not null}
		\State $\textproc{push}(\textproc{GetSubsection}(\monoSection(t), t^\downarrow)) \rightarrow \monoSects$
		\EndIf
		\EndWhile
		\State \textbf{return} true
		\EndFunction
		
	\end{algorithmic}
\end{algorithm}

We first split the trajectory $\camPos(t)$ into sections with monotonically changing depth as described in Section \ref{sec:monotonic}, and insert the sections into list $\monoSects$ using $\textproc{GetMonotonicSections}$ (line 4).
Then, for each monotonic section $\monoSection(t)$, we compute the deepest point $\deepestPoint$ (i.e. one of the endpoints of the section), and try to find a pyramid containing that point (line 6-8).
The function $\textproc{FindContainingPyramid}$ (line 8) returns either a pyramid that contains $\deepestPoint$ or null, indicating no pyramid containing $\deepestPoint$ was found.
If no pyramid in $\pyramidSet$ contains $\deepestPoint$, we attempt to generate a new pyramid using the method described in the previous subsection (line 10), but if pyramid generation fails we declare the trajectory to be in collision (line 12).

Next, we try to compute the deepest point at which the monotonic section $\monoSection(t)$ intersects one of the four lateral faces of the pyramid $\pyramid$.
Using the method described in Section \ref{sec:intersections}, we compute the times at which $\monoSection(t)$ intersects each lateral face of the pyramid, and choose the time $\intersectionTime$ at which $\monoSection(t)$ has the greatest depth (line 14).
If $\monoSection(t)$ is found to not collide with any of the lateral faces of the pyramid, then it necessarily must remain inside the pyramid and the section can be declared collision-free.
However, if $\monoSection(t)$ does collide with one of the lateral faces of the pyramid, we split it at $\intersectionTime$ and add the section of $\monoSection(t)$ that is outside of the pyramid to $\monoSects$ (line 16).
Thus, if each subsection of the trajectory is found to be inside the space covered by the set of pyramids $\pyramidSet$, then the trajectory is declared collision-free (line 17).

Note that this method of collision checking allows for pyramids to be generated on an as-needed basis rather than requiring all pyramids to be generated in a batch process when a new depth image arrives.
This additionally avoids generating unneeded pyramids; only those required for collision checking are created. 

\subsection{Planning algorithm}\label{sec:planningAlg}
Algorithm \ref{algo:planner} describes the path planning algorithm used in this work.
The algorithm takes as input the most recently received depth image and vehicle state estimate, where the state estimate partially defines each candidate trajectory as given in \eqref{eq:camTrajDef}.
Within a user-specified time budget, the algorithm repeatedly generates and evaluates candidate trajectories for cost and the satisfaction of constraints, returning the lowest cost trajectory that satisfies all constraints.
We choose to use a random search algorithm due to its simplicity and probabilistic optimality, though the collision checking algorithm presented in the previous subsection can be used in conjunction with other planning algorithms as well (see \cite{karaman2011sampling}, for example).

\begin{algorithm}
	\caption{Lowest Cost Trajectory Search}
	\label{algo:planner}
	\begin{algorithmic}[1]
		\State \textbf{input:} Latest depth image $\depthImage$ and vehicle state
		\State \textbf{output:} Lowest cost collision-free trajectory found $\bestCamPos(t)$ or an undefined trajectory (indicating failure)
		
		\Function{FindLowestCostTrajectory()}{}
		\State $\bestCamPos(t) \gets$ undefined with $\textproc{Cost}(\bestCamPos(t)) = \infty$
		\State $\pyramidSet \gets \emptyset$
		\While{computation time not exceeded}
		\State $\camPos(t) \gets \textproc{GetNextCandidateTraj}()$
		\If{$\textproc{Cost}(\camPos(t)) < \textproc{Cost}(\bestCamPos(t))$}
		\If{\textproc{IsDynamicallyFeas}($\camPos(t)$)}
		\If{\textproc{IsCollisionFree}($\camPos(t)$, $\pyramidSet$, $\depthImage$)}
		\State $\bestCamPos(t) \gets \camPos(t)$
		\EndIf
		\EndIf
		\EndIf
		\EndWhile
		\State \textbf{return} $\bestCamPos(t)$
		\EndFunction
		
	\end{algorithmic}
\end{algorithm}

Let $\textproc{GetNextCandidateTraj}$ be defined as a function that returns a randomly generated candidate trajectory $\camPos(t)$ using the methods described in \cite{mueller2015computationally} (line 7).
The function $\textproc{Cost}$ is a user-specified function used to compare candidate trajectories (line 8).
In this work, we define $\textproc{Cost}$ to be the following, where $\expDir$ is a desired exploration direction: 
\begin{equation}\label{eq:cost}
 \textproc{Cost}(\camPos(t)) = \frac{\expDir \cdot \left(\camPos(0) - \camPos(\finalTime)\right)}{\finalTime}
\end{equation}
That is, better trajectories are those that cause the vehicle to move quickly in the desired direction $\expDir$.
Note, however, that $\textproc{Cost}$ can be defined arbitrarily by the user to include other objectives based on the desired behavior of the vehicle (e.g. to favor increased distance to other vehicles or people).

The function $\textproc{IsDynamicallyFeas}$ (line 9) checks whether the given candidate trajectory satisfies constraints on the total thrust and angular velocity of the vehicle using methods described in \cite{mueller2015computationally}.
Finally, the candidate trajectory is checked for collisions with the environment using $\textproc{IsCollisionFree}$ (line 10).
We check for collisions with the environment last because it is the most computationally demanding step of the process.

In this way, Algorithm \ref{algo:planner} can be used as a high-rate local planner that ensures the vehicle avoids obstacles, while a global planner that may require significantly more computation time can be used to specify high-level goals (e.g. the exploration direction $\expDir$) without the need to worry about obstacle avoidance or respecting the dynamics of the vehicle.
We run Algorithm \ref{algo:planner} in a receding-horizon fashion, where each new depth image is used to compute a new collision-free trajectory.
We additionally constrain our candidate trajectories to bring the vehicle to rest, so that if no feasible trajectories are found during a given planning step, the last feasible trajectory can be tracked until the vehicle comes to rest.

\section{Algorithm Performance} \label{sec:algBenchmark}

In this section we assess the performance of the proposed algorithm in terms of its conservativeness in labeling trajectories as collision-free, its speed, and its ability to evaluate a dense set of candidate trajectories in various amounts of compute time.
We additionally compare our method to other state-of-the-art memoryless planning algorithms.

\begin{figure}
	\includegraphics[width=\columnwidth]{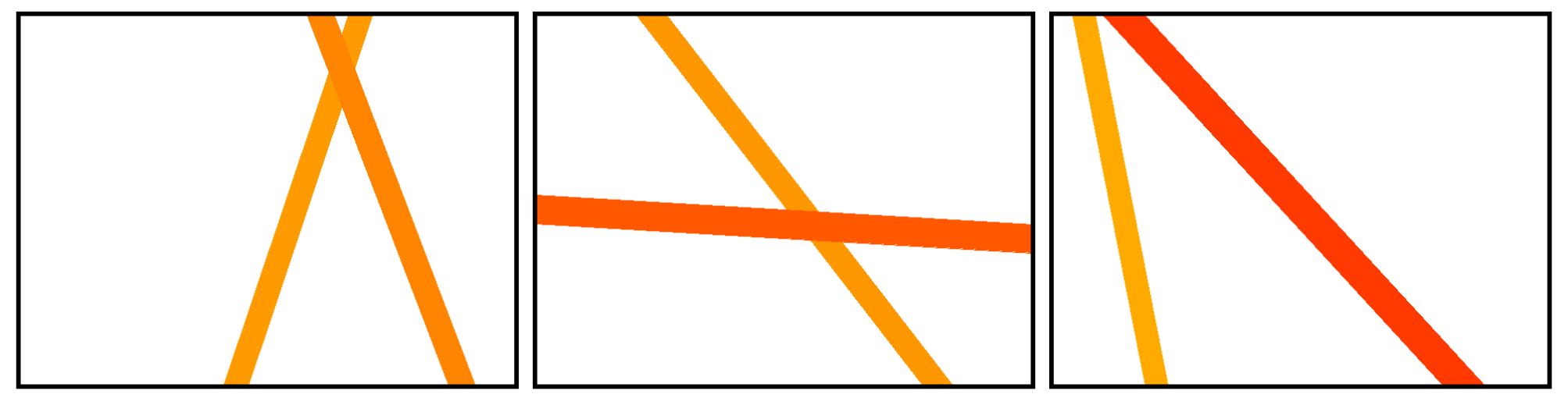}
	\centering
	\vspace{-0.5cm}
	\caption{Three examples of synthetic depth images used for benchmarking the proposed algorithm. Two rectangular obstacles are generated at different constant depths. The background is considered to be at infinite depth.}
	\label{fig:benchmark}
\end{figure}

To benchmark our collision checking method, we conduct various Monte Carlo simulations using a series of randomly generated synthetic depth images and vehicle states.
Examples of several generated depth images are shown in Figure \ref{fig:benchmark}.
The image is generated by placing two \SI{20}{\centi\meter} thick rectangles with random orientations in front of the camera at distances sampled uniformly at random on (\SI{1.5}{\meter}, \SI{3}{\meter}).
Note that this choice of obstacles is arbitrary; any number, distribution, or type of obstacles could be used to conduct such tests.
However, rather than trying to emulate a specific type of environment, we choose to use obstacles in our benchmark that are primarily easy to both visualize and reason about conceptually.
Furthermore, the use of such obstacles does not unfairly benefit the proposed collision checking method by, e.g., breaking the free-space into regions that may be easier to describe using pyramids.

The initial velocity of the vehicle in the camera-fixed z-direction $\camFrameZ$ is sampled uniformly on  (\SI{0}{\meter\per\second}, \SI{4}{\meter\per\second}), and the initial velocity of the vehicle in both the x-direction $\camFrameX$ and y-direction $\camFrameY$ is sampled uniformly on (\SI{-1}{\meter\per\second}, \SI{1}{\meter\per\second}).
We assume the camera is mounted such that $\camFrameZ$ is perpendicular to the thrust direction of the vehicle, and thus set the initial acceleration of the vehicle in both the $\camFrameX$ and $\camFrameZ$ directions to zero.
The initial acceleration in the $\camFrameY$ direction is sampled uniformly on (\SI{-5}{\meter\per\second\squared}, \SI{5}{\meter\per\second\squared}).

The planner generates candidate trajectories that come to rest at randomly sampled positions in the field of view of the depth camera.
Specifically, positions in the depth image are sampled uniformly in pixel coordinates and then deprojected to a depth that is sampled uniformly on (\SI{1.5}{\meter}, \SI{3}{\meter}).
The duration of each trajectory is sampled uniformly on (\SI{2}{\second}, \SI{3}{\second}).

The algorithm was implemented in C++ and compiled using GCC version 5.4.0 with the -O3 optimization setting.
Three platforms were used to assess performance: a laptop with an Intel i7-8550U processor set to performance mode, a Jetson TX2, and an ODROID-XU4.
The algorithm is run as a single thread in all scenarios.

\subsection{Conservativeness}
We first analyze the accuracy of the collision checking method described by Algorithm \ref{algo:collisionCheck}.
A key property of our method is that it will never erroneously label a trajectory as collision-free that either collides with an obstacle or has the potential to collide with an occluded obstacle.
Such a property is typically a requirement for collision checking algorithms used with aerial vehicles, as a trajectory mislabeled as collision-free can result in a catastrophic crash resulting in a total system failure.

However, because the generated pyramids cannot exactly describe the free space of the environment, our method may erroneously label some collision-free trajectories as being in-collision.
In order to quantify this conservativeness, we compare our method to a ground-truth, ray-tracing based collision checking method capable of considering both field-of-view constraints and occluded obstacles.
We define conservativeness as the number of trajectories erroneously labeled as in-collision divided by the total number of trajectories labeled (both correctly and incorrectly) as in-collision.
A single test consists of first generating a synthetic scene and random initial state of the vehicle as previously described.
We then generate 1000 random trajectories for each scene, and perform collision checking both with our method and the ground-truth method.
The number of trajectories both correctly and incorrectly labeled as in-collision are averaged over $10^4$ such scenes.
Additionally, in order to quantify how well the environment can be described using the pyramids generated by our method, we limit the number of pyramids the collision checking algorithm is allowed to use, and repeat this test for various pyramid limits.

\begin{figure}
	\includegraphics[width=\columnwidth]{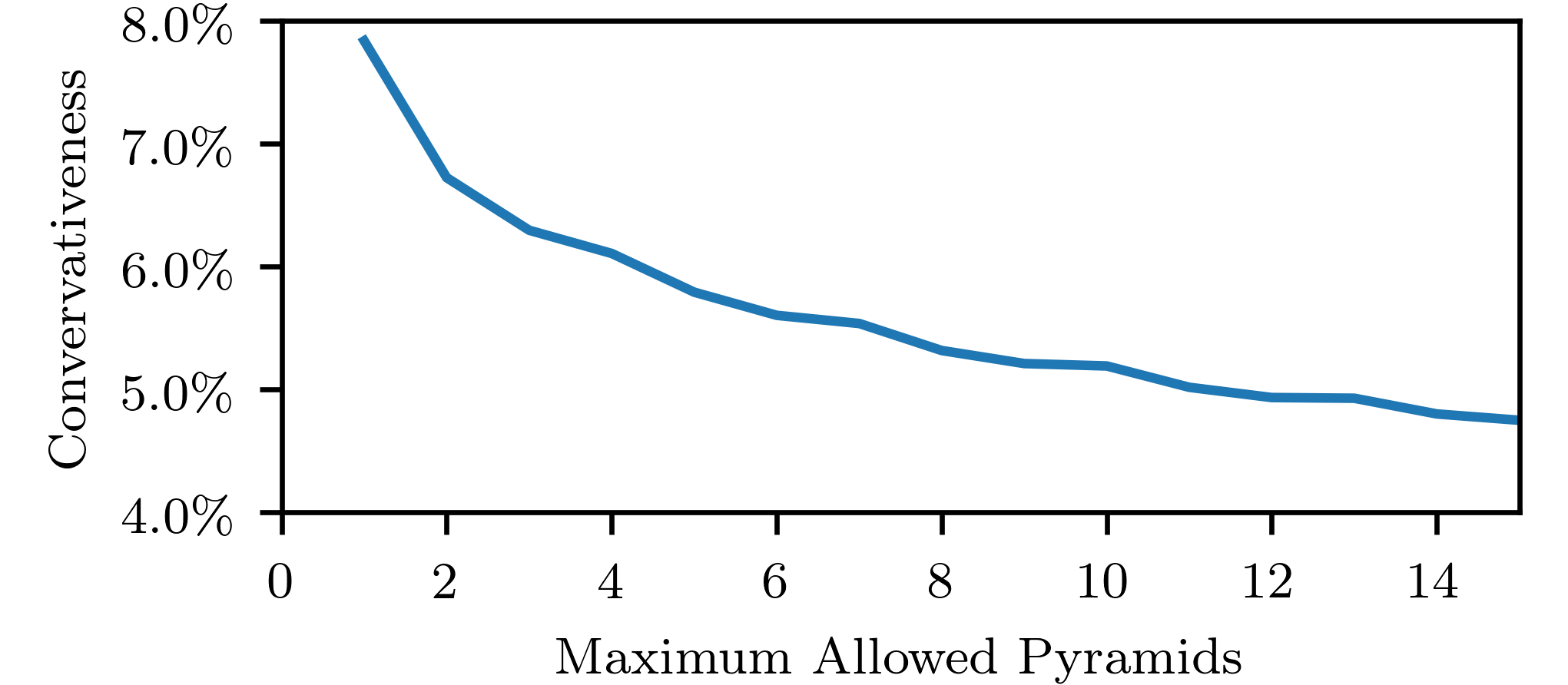}
	\centering
	\caption{Conservativeness of the collision checking algorithm as a function of the maximum number of pyramids allowed to be generated. We define conservativeness as the number of trajectories erroneously labeled as in-collision divided by the total number of trajectories labeled as in-collision. The free-space of the environment is described with increasing detail as more pyramids are allowed to be generated, leading to a lower number of trajectories being erroneously labeled as in-collision.}
	\label{fig:correctness}
\end{figure}

Figure \ref{fig:correctness} shows how the over-conservativeness of our method decreases as the number of pyramids allowed to be used for collision checking increases.
The percent of mislabeled trajectories is initially higher because the environment cannot be described with high accuracy using fewer pyramids.
However, conservativeness remains nearly constant for larger pyramid limits, indicating that our method may erroneously mislabel a small number of collision-free trajectories (e.g. those in close proximity to obstacles).
Note that we do not limit the number of pyramids generated when using the planning algorithm described in Algorithm \ref{algo:planner}, as we have found it to be unnecessary in practice.

\subsection{Collision Checking Speed}
Next we compare our collision checking method to the state-of-the-art k-d tree based methods described in \cite{florence2016integrated} and \cite{lopez2017aggressive}.
Both our method and k-d tree methods involve two major steps: the building of data structures (i.e. a k-d tree, or the pyramids described in this paper) and the use of those data structures to perform collision checking with the environment.
Our method differs from k-d tree based methods, however, in its over-conservativeness.
Specifically, we consider trajectories that pass through occluded space to be in collision with the environment, while k-d tree based methods only consider trajectories that pass within the vehicle radius of detected obstacles to be in collision.

We compare our method to the k-d tree methods by first limiting the amount of time allocated for pyramid generation such that it is similar to the time required to build a k-d tree as reported in \cite{florence2016integrated} and \cite{lopez2017aggressive} (roughly \SI{1.81}{\milli\second}).
We then check 1000 trajectories for collisions, and compute the average time required to check a single trajectory for collisions using the generated pyramids.
Similar to \cite{florence2016integrated} and \cite{lopez2017aggressive}, we use a $160\times 120$ resolution depth image which we generate using the previously described method, and average our results over $10^{4}$ Monte Carlo trails.

Table \ref{tab:benchComparison} shows how the average performance of our method outperforms the best-case results reported by \cite{florence2016integrated} and \cite{lopez2017aggressive}.
On average 27.5, 19.3, and 15.3 pyramids were generated during the allocated \SI{1.81}{\milli\second} pyramid generation time on i7, TX2, and ODROID platforms respectively.
The difference in computation time can be reasoned about using a time complexity analysis.
Let a given depth image contain $\numPixels$ pixels.
Then $O(\numPixels \mathrm{log}(\numPixels))$ operations are required to build a k-d tree, while $O(\numPixels)$ operations are required to generate a single pyramid (of which there are typically few).
Because a single k-d tree query takes $O(\mathrm{log}(\numPixels))$ time, if the trajectory must be checked for collisions at $m$ sample points along the trajectory, then the time complexity for checking a single trajectory for collisions is $O(m\mathrm{log}(\numPixels))$.
However, collision checking a single trajectory using our method can be done in near constant time, as it only requires finding the roots of several fourth order polynomials (which is done in closed-form) as described in Section \ref{sec:intersections}.
Additionally, note that while an entire k-d tree must be built before being used to check trajectories for collisions, the pyramids generated by our method can be built on an as-needed basis, reducing computation time even further.

\begin{table}[]
	\centering
	\caption{Average Collision Checking Time Per Trajectory}
	\label{tab:benchComparison}
	\begin{tabular}{|l|c|c|}
		\hline
		& \multicolumn{1}{|c|}{\begin{tabular}[c]{@{}c@{}} Computer \\ \end{tabular}} & \multicolumn{1}{|c|}{\begin{tabular}[c]{@{}c@{}}Single Trajectory \\ Collision Check (\SI{}{\micro\second}) \end{tabular}} \\ \hline
		Florence et al.\footnotemark \cite{florence2016integrated} & i7 NUC & 56 \\ \hline
		Lopez et al.\footref{foot:refnote} \cite{lopez2017aggressive} & i7-2620M & 48 \\ \hline
		RAPPIDS (ours) & i7-8550U & 1.20 \\ \hline
		RAPPIDS (ours) & Jetson TX2 & 3.81 \\ \hline
		RAPPIDS (ours) & ODROID-XU4 & 8.72 \\ \hline
	\end{tabular}
\end{table}
\footnotetext{The best reported average collision checking time required per trajectory is used for comparison.\label{foot:refnote}}

\subsection{Overall Planner Performance}
Finally, we describe the overall performance of the planner, i.e. Algorithm \ref{algo:planner}, using the same Monte Carlo simulation but with $640\times 480$ resolution depth images, which are the same resolution as those used in the physical experiments described in the following section.
The number of trajectories evaluated by the planner is used as a metric of performance, where a larger number of generated trajectories indicates a better coverage of the trajectory search space and thus higher likelihood of finding the lowest possible cost trajectory within the allocated planning time.

\begin{figure}
	\includegraphics[width=\columnwidth]{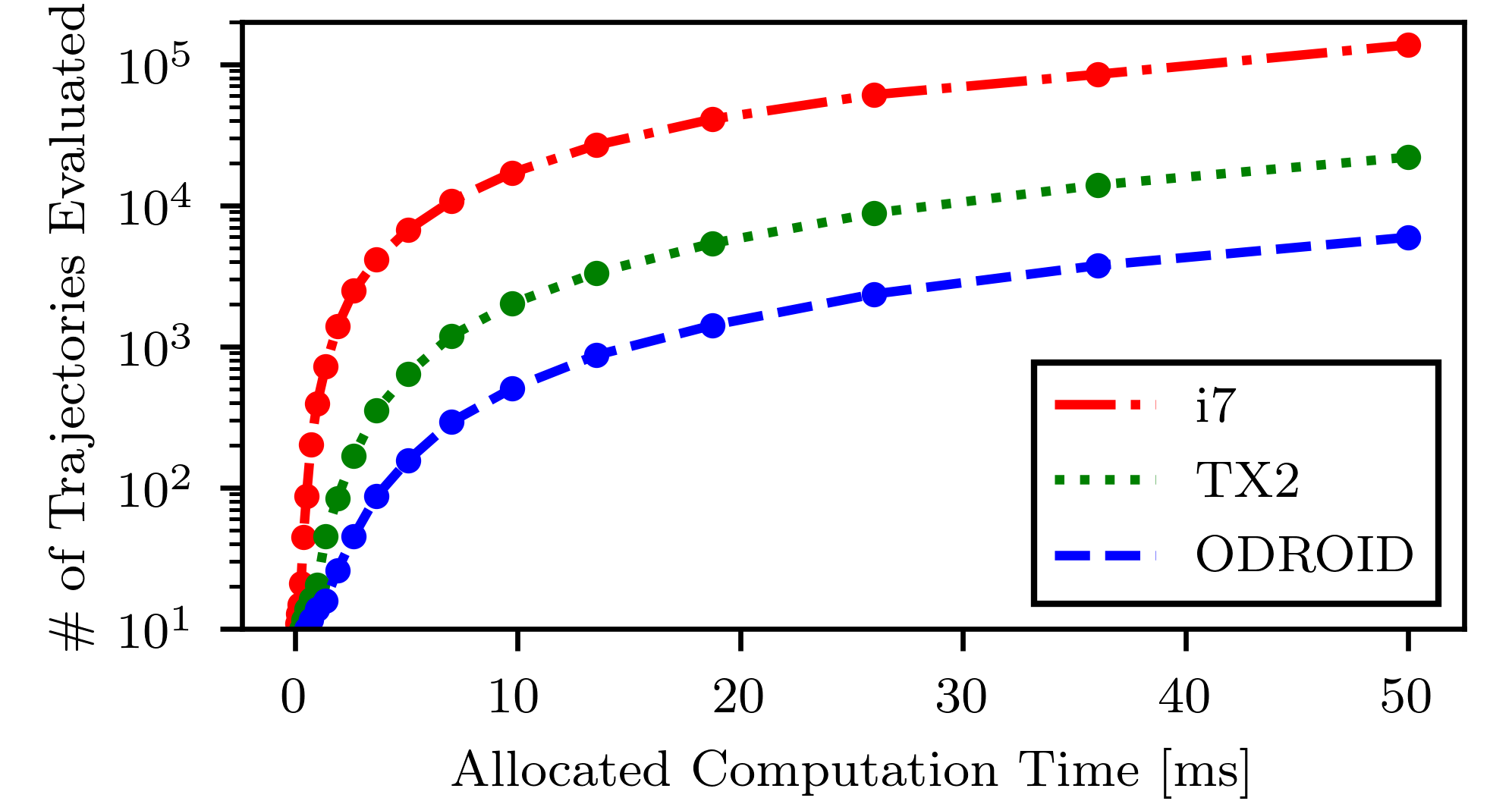}
	\centering
	\caption{Average planner performance as a function of allocated computation time across various platforms. As computation time increases, the number of trajectories evaluated increases at different rates for platforms with different amounts of computation power.}
	\label{fig:avgTrajGen}
\end{figure}

Figure \ref{fig:avgTrajGen} shows the results of running the planner for $10^4$ Monte Carlo trails each on the i7-8550U processor, the Jetson TX2, and the ODROID-XU4 for computation time limits between \SI{0}{\milli\second} and \SI{50}{\milli\second}.
Naturally, as computation time increases, the average number of trajectories evaluated increases monotonically.
Furthermore, we observe that the i7-8550U outperforms the Jetston TX2, which outperforms the ODROID-XU4.
The difference in performance can be explained by the fact that the Jetston TX2 and especially the ODROID-XU4 are intended to be low-power devices capable of being used in embedded applications.
However, due to the computational efficiency of our collision checking method, we found that even the ODROID-XU4 is capable of evaluating a sufficiently large number of trajectories within a small amount of time.
This makes it feasible to use low-power devices such as the ODROID-XU4 as onboard flight controllers while still achieving fast, reactive flight.

\section{Experimental Results} \label{sec:experiment}

\begin{figure}
	\includegraphics[width=\columnwidth]{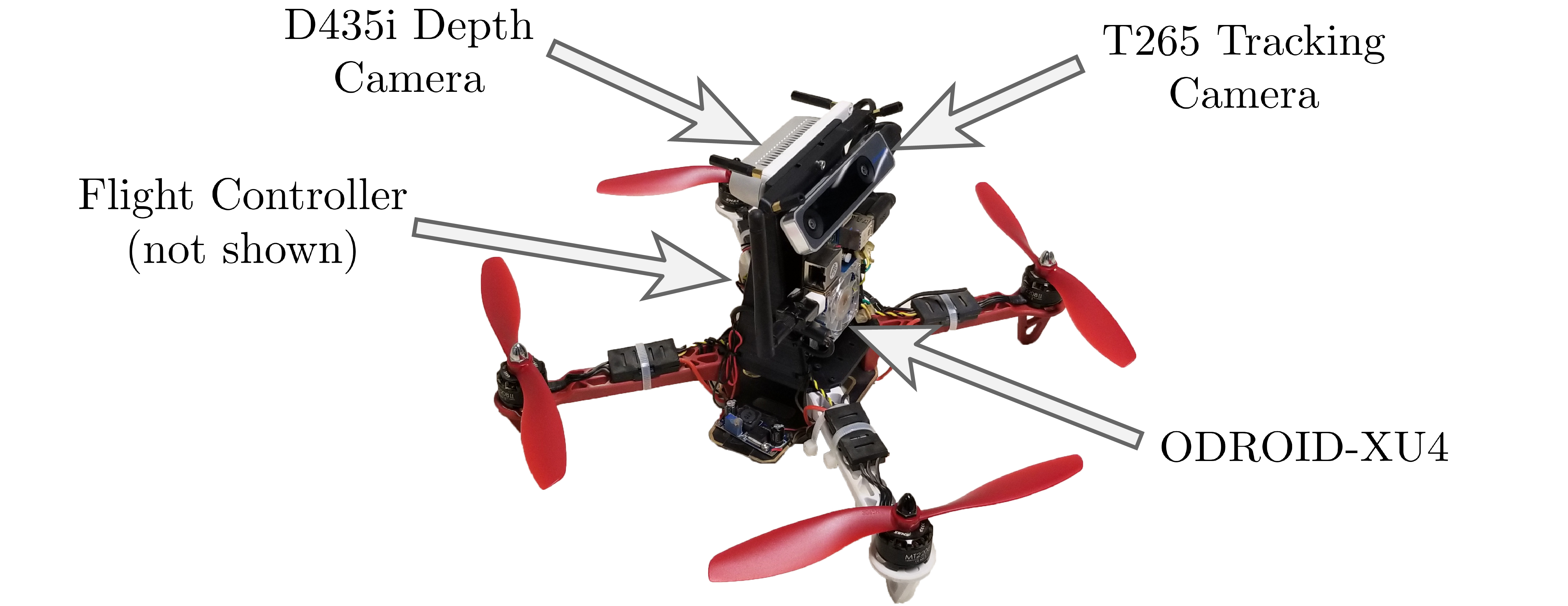}
	\centering
	\caption{Vehicle used in experiments. A RealSense D435i depth camera is used to acquire depth images, and a RealSense T265 tracking camera is used to obtain state estimates of the vehicle. The proposed algorithm is run on an ODROID-XU4, which sends desired thrust and angular velocity commands to a Crazyflie 2.0 flight controller.}
	\label{fig:vehicle}
\end{figure}

In this section we demonstrate the use of the proposed algorithm on an experimental quadcopter, shown in Figure \ref{fig:vehicle}.
The quadcopter has a mass of \SI{1.0}{\kilo\gram}, and is equipped with an ODROID-XU4, RealSense D435i depth camera, RealSense T265 tracking camera, and Crazyflie 2.0 flight controller.
The ODROID is used in order to demonstrate the feasibility of running the proposed algorithm at high rates on cheap, low mass, and low power hardware.
The tracking camera provides pose estimates to the ODROID at \SI{200}{\hertz}, which a Kalman filter uses to compute translational velocity estimates.
Filtered acceleration estimates are obtained at \SI{100}{\hertz} using the IMU onboard the crazyflie flight controller.
The depth camera is configured to capture $640 \times 480$ resolution depth images at \SI{30}{\hertz}, and the proposed planning algorithm is run for \SI{30}{\milli\second} when each new depth image arrives using the latest state estimate provided by the Kalman filter.
If no collision-free trajectories can be found during a given planning stage, the vehicle continues to track the most recently found collision-free trajectory from a previous planning stage until either a new collision-free trajectory is found or the vehicle comes to rest.

\begin{figure}
	\includegraphics[width=\columnwidth]{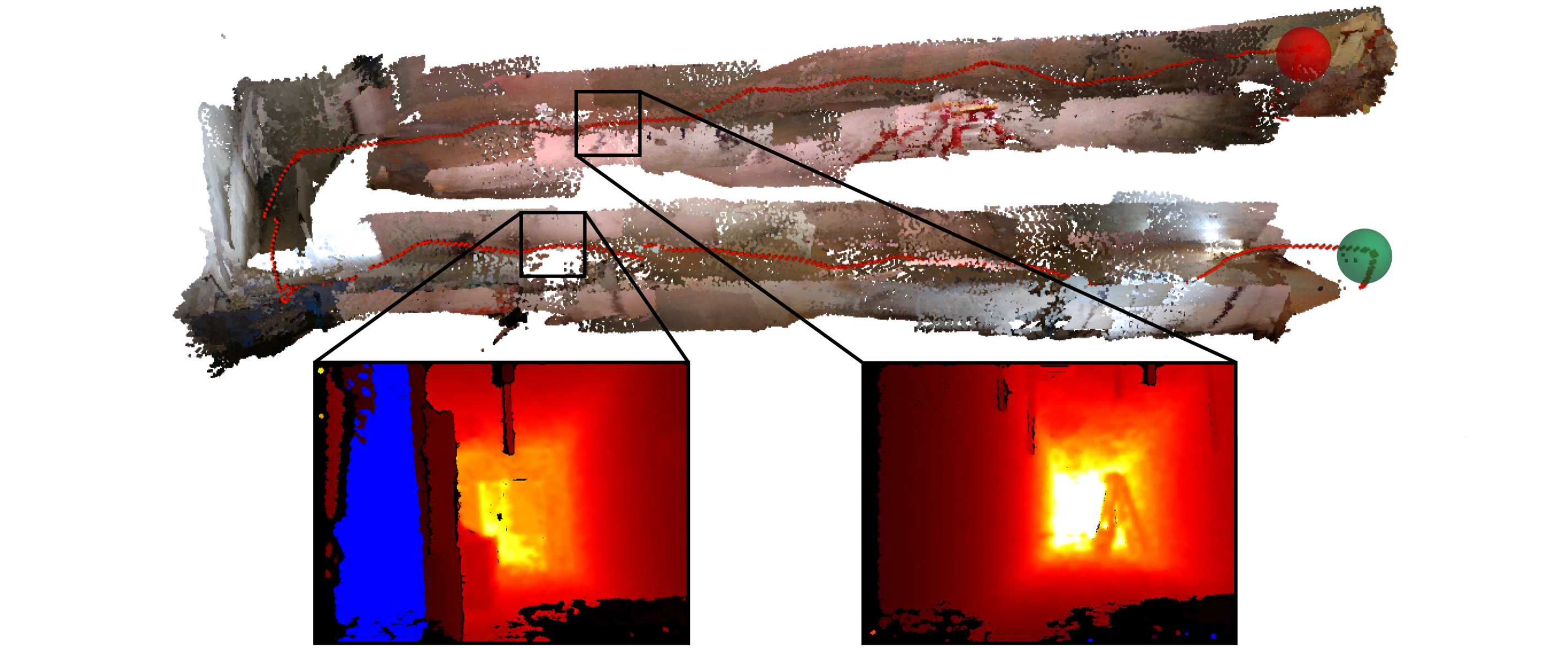}
	\centering
	\caption{Visualization of flight experiment in U-shaped tunnel environment. The path of the vehicle is shown as a red line. The vehicle starts at the green sphere and ends at the red sphere. The map of the environment (top) is generated at the end of the experiment using depth images captured by the depth camera. Two depth images (bottom) where no collision-free trajectories were found are shown to illustrate cases where the planner fails. Pixels with depth values less than \SI{0.75}{\meter} but greater than the vehicle radius are highlighted in blue. In the left image, an obstacle occludes a significant portion of the image, preventing collision-free trajectories from being found due to the proximity of the obstacle to the vehicle. In the right image, a very small amount of noise is present near the bottom of the image, causing the planner to hallucinate the presence of close proximity obstacles in what would otherwise be free-space.	A full video of the experiment is attached to this paper.\protect\footref{foot:youtube}}
	\label{fig:tunnelTraj}
\end{figure}

The vehicle was commanded to fly in a U-shaped tunnel environment that was previously unseen by the vehicle, shown in Figure \ref{fig:tunnelTraj}.
Each branch of the tunnel measured roughly \SI{2.5}{\meter} in width and height, \SI{20}{\meter} in length, and was filled with various obstacles for the vehicle to avoid.
The candidate trajectories generated by the planner were generated using the same method described in Section \ref{sec:algBenchmark}.
A video of the experiment is attached to this paper.\footnote{The video can also be viewed at \url{https://youtu.be/Pp-HIT9S6ao} \label{foot:youtube}}

The desired exploration direction $\expDir$ used to compute the cost of each candidate trajectory as given by \eqref{eq:cost} is set as follows.
We initialize $\expDir$ to be horizontal and to point down the first hallway.
When the vehicle is at rest and no feasible trajectories are found by the planner, the desired exploration direction $\expDir$ is rotated \SI{90}{\degree} to the right of the vehicle, allowing the vehicle to navigate around corners.
We then stop the test when the vehicle reaches the end of the second hallway.
We use this method of choosing the exploration direction simply as a matter of convenience in demonstrating the use of our algorithm in a cluttered environment.
However, many other suitable methods of providing high-level goals to our algorithm can be used (e.g. \cite{cieslewski2017rapid}), but are typically application dependent and thus are not discussed here.

During the experiment, the vehicle was able to find at least one collision-free trajectory in 35.3\% of the \SI{30}{\milli\second} planning stages.
Of the planning stages where at least one feasible trajectory was found, 2069.2 candidate trajectories and 2.9 pyramids were generated on average.
The vehicle traveled approximately \SI{40}{\meter} over \SI{43}{\second}, and attained a maximum speed of \SI{2.66}{\meter\per\second}.

\begin{figure}
	\includegraphics[width=\columnwidth]{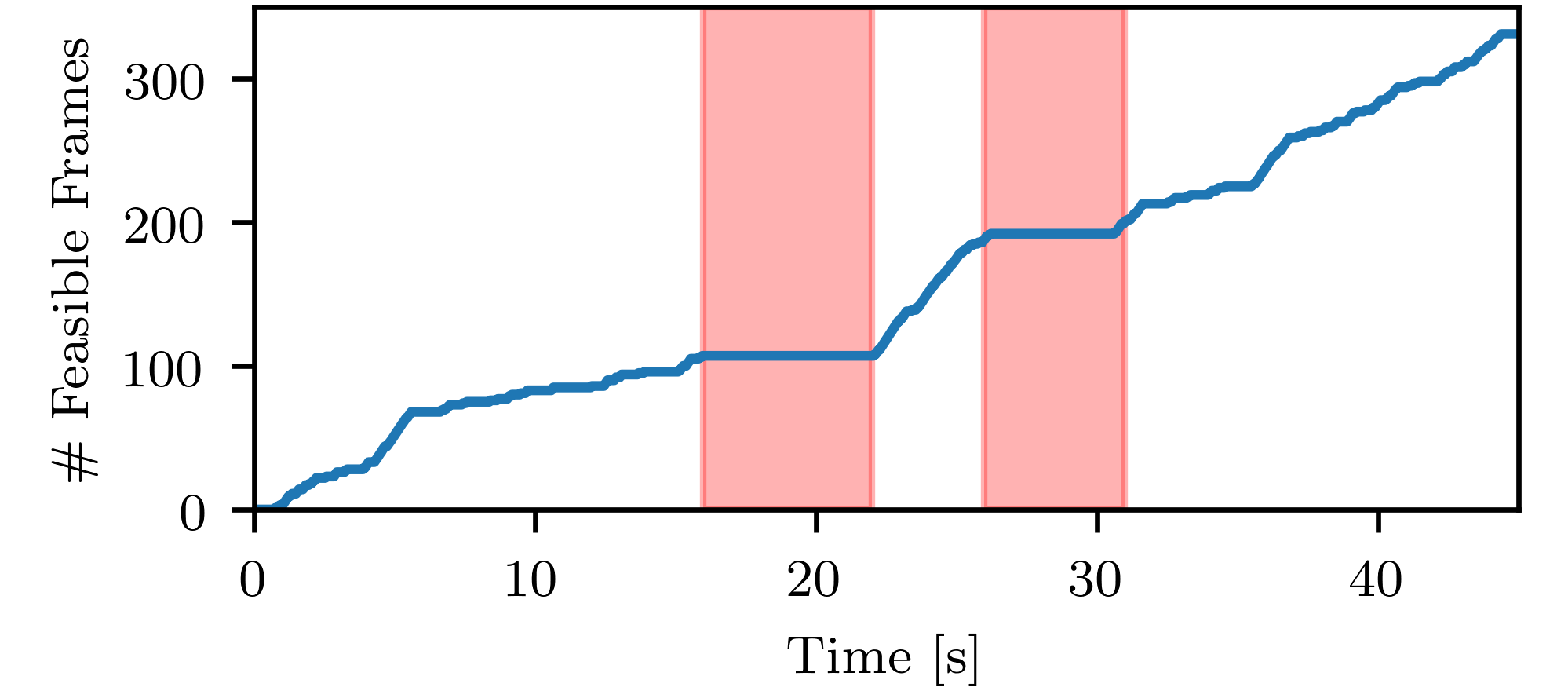}
	\centering
	\caption{Cumulative number of planning stages where at least one collision-free trajectory was found. The sections of the graph highlighted in red correspond to periods in which the vehicle is facing the wall at the end of the hallway. A large increase in successful planning stages is observed between \SI{22}{\second} and \SI{26}{\second} when the vehicle is navigating in the relatively uncluttered area between the two hallways.}
	\label{fig:feasResults}
\end{figure}

The low percentage of planning stages where at least one collision-free trajectory was found primarily are cases where the vehicle passes closely to obstacles and also by the significant amount of noise present in the depth images.
Figure \ref{fig:tunnelTraj} shows examples of both cases.
Note that the amount of noise present in the depth images can be reduced via filtering, although this may lead to the potential misdetection of small and/or thin obstacles.
Additionally, Figure \ref{fig:feasResults} shows how the successful planning stages are distributed throughout the experiment.
A lower percent of successful planning stages is observed when the vehicle is navigating the cluttered hallways than the relatively open area between the two hallways, which is potentially due to the difference in obstacle density and lighting conditions (leading to a difference in depth image noise levels).

\section{Conclusion} \label{sec:conclusion}

In this paper we presented a novel pyramid-based spatial partitioning method that allows for efficient collision checking between a given trajectory and the environment as represented by a depth image.
The method allows multicopters with limited computational resources to quickly navigate cluttered environments by generating collision-free trajectories at high rates.
A comparison to existing state-of-the-art depth-image-based path planning methods was performed via Monte Carlo simulation, showing our method to significantly reduce the computation time required to perform collision checking with the environment while being more conservative than other methods by implicitly considering occluded obstacles.
Finally, real-world experiments were presented that demonstrate the use of our algorithm on computationally low-power hardware to perform fully autonomous flight in a previously unseen, cluttered environment.

\section*{Acknowledgement}
This material is based upon work supported by the Berkeley Fellowship for Graduate Study, the Berkeley DeepDrive project `Autonomous Aerial Robots in Dense Urban Environments', and the China High-Speed Railway Technology Co., Ltd.
The experimental testbed at the HiPeRLab is the result of contributions of many people, a full list of which can be found at \url{hiperlab.berkeley.edu/members/}.

\bibliographystyle{IEEEtran}
\bibliography{references}

\end{document}